\documentclass[twoside,11pt]{article}

\usepackage{jmlr2e}
\usepackage{algorithmicx}

\usepackage{caption}
\usepackage{subcaption}

\usepackage{algorithm}
\usepackage{algpseudocode}

\jmlrheading{1}{}{}{}{}{}

\firstpageno{1}

\begin{document}

\title{Deep Surrogate of Modular Multi Pump using Active Learning}

\author{\name \emph{Malathi} Murugesan \email malathi.murugesan@bakerhughes.com \\
\name \emph{Kanika} Goyal \email kanika.goyal@bakerhughes.com \\
\name \emph{Laure} Barriere \email laure.barriere@bakerhughes.com \\
\name \emph{Maura} Pasquotti \email maura.pasquotti@bakerhughes.com \\
\name \emph{Giacomo} Veneri \email giacomo.veneri@bakerhughes.com \\
\name \emph{Giovanni} De Magistris \email giovanni.demagistris@bakerhughes.com \\
\addr AI Team - Baker Hughes, Florence, Italy \and Le Creusot, France \and Bangalore, India }


\maketitle

\begin{abstract}
Due to the high cost and reliability of sensors, the designers of a pump reduce the needed number of sensors for the estimation of the feasible operating point as much as possible. The major challenge to obtain a good estimation is the low amount of data available. Using this amount of data, the performance of the estimation method is not enough to satisfy the client requests.  
To solve this problem of scarcity of data, getting high quality data is important to obtain a good estimation. Based on these considerations, we develop an active learning framework for estimating the operating point of a Modular Multi Pump used in energy field. In particular we focus on the estimation of the surge distance.
We apply Active learning to estimate the surge distance with minimal dataset. Results report that active learning is a valuable technique also for real application.
\end{abstract}


\section{Introduction}
\begin{figure}
     \centering
     \begin{subfigure}[b]{0.3\textwidth}
         \centering
         \includegraphics[width=\textwidth]{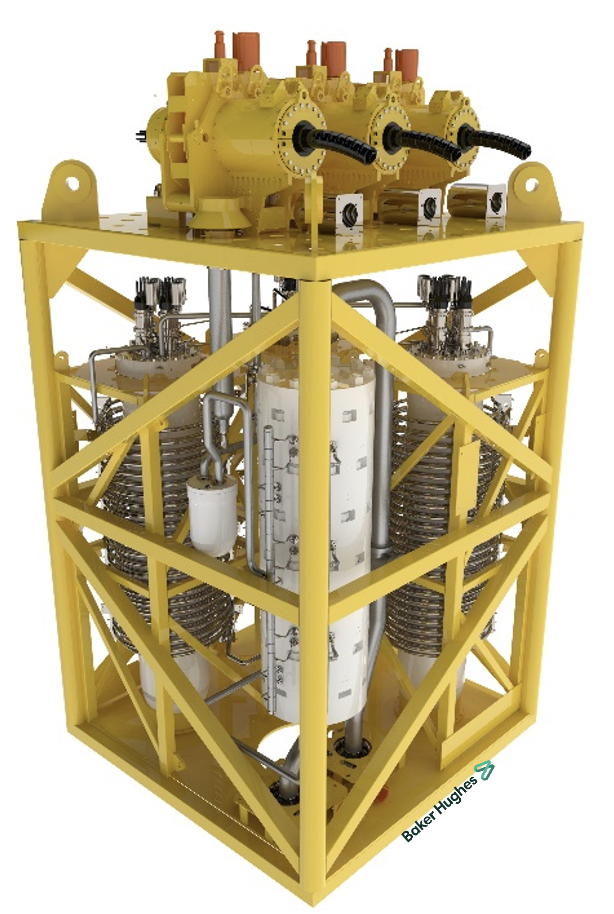}
         \caption{12-stage pump module.}
         \label{fig:photo}
     \end{subfigure}
     \hfill
     \begin{subfigure}[b]{0.5\textwidth}
         \centering
         \includegraphics[width=\textwidth]{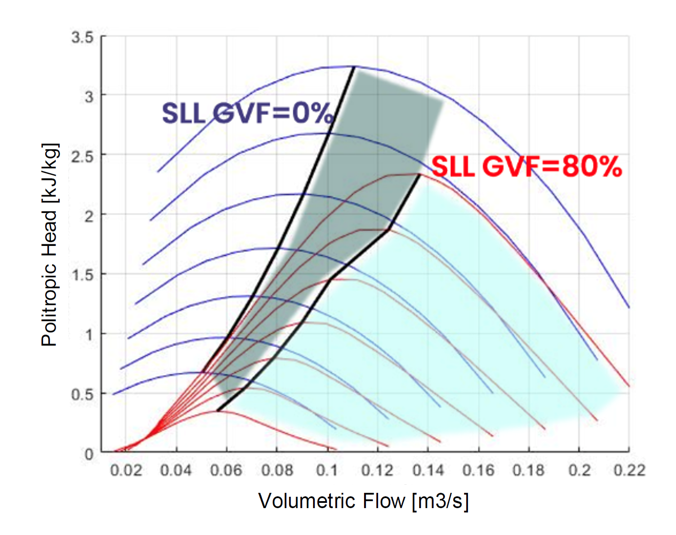}
         \caption{Distance of operating point of the compressor from the surge limit line (SLL)}
         \label{fig:surge}
     \end{subfigure}
        \caption{Pump module and surge phenomenon.}
        \label{fig:three graphs}
\end{figure}

In subsea energy field, we often apply boosting solutions to increase the pressure of the fluid extracted 
to overcome the continuous pressure decline that affects the real production. Common best practices in presence of multiphase fluid are based on the combination of a separator and a single-phase compressor underwater. To reduce complexity of the equipment, reduce footprint and simplify connections, a multiphase pump (Figure \ref{fig:photo}) able to directly process and compress the liquid and gas mixture has been identified. To control the multiphase pump, the controller needs to know its operating point, in terms of Volumetric inlet flow ($Q_{in}$) processed, Gas Volume Fraction ($\textnormal{GVF}$) and fluid composition. The estimation of these parameters allows the calculation of the \textit{surge distance} and then to control the pump desired behaviors. The surge distance is the difference of the compressor operating point from the chosen Surge Limit Line (SLL) (Fig. \ref{fig:surge}). The controller of the pump needs to keep the operating point 5-10\% distant from SLL to have enough time to recover. The estimation of the surge distance reduces the risk for the pump of the “Surge” phenomenon \cite{Hafaifa2014}\cite{SEMLITSCH2016}. This phenomenon is an operating point where downstream pressure of pump discharge rapidly increases due to a decrease in flow demand from the pump outlet.

The problem of estimating the operating point for multi pump systems is challenging because the number of sensors is limited. This is due to the high cost of sensors
and strict requirements for maintenance minimization of the system deployed on the seabed. Therefore, only a limited subset of measurements is available, while all the others need to be properly estimated. 

In order to estimate the operating point 
and avoid the surge phenomenon, a first solution is to use physical models (PHM) in the design of the pump. However, brute force design optimization using PHM is an intensive and long running process. Multi pumps have an increasing number of design parameters and looking for best ones 
(using brute-force simulations) becomes increasingly expensive. 
For this reason, we developed a surrogate model able to estimate the operating point of the modular multi pump. To obtain an accurate model from minimal training data, we developed an active learning approach \cite{markus2013,Shubhomoy2016,settles2009active} in which training points are selected based on error measure as well as uncertainty measure through which we can reduce the number of training points by more than an order of magnitude for a neural network surrogate model \cite{xiang2020active,Boyd2004, bradley2021, heavlin2021}.
Moreover, performing real tests is 
a demanding and costly process and we need to optimize our Design of Experiment (DoE) \cite{Palermo2009}.



\subsection{Related Works}
\small
Active learning has been extensively used for experiment design and optimization \cite{Yu2006, settles2009active, markus2013, Shubhomoy2016, Jamieson2015, bradley2021}. Indeed, \citet{Yu2006} has extensively demonstrated the efficiency of active learning to select the most informative experiment on synthetic data; while,  \citet{Jamieson2015} proposed a framework to work with real application and to collect large-scale datasets. Different techniques have been proposed to apply active learning to real application (see \cite{Sverchkov2017} for a review), suggesting one optimal strategy \cite{bradley2021} to maximize the benefit of getting new experiment.

In this paper we aim to apply active learning methodology to improve design capabilities of pump and compressors.
To validate the active learning methodology, we used datasets generated by PHM and we limited to a realistic case (no more than 10\% samples).

\section{Dataset}
\small

The dataset used to validate the approach is a Design of Experiment produced by running the design tool PHM. Data are produced for the first stage only. Data cover extensively the parameter space. For instance, inlet temperature takes all the values between 293.15 K and 353.15 K. We analyze the following data (see Table \ref{tab:symbols}): 
- Input data: inlet temperature ($Tin$), inlet pressure ($Pin$), rotational speed ($N$), differential pressure ($\Delta P = Pin - Pout$) and total consumed power ($Power$) - Output data: \textbf{surge distance ($SD$)} - Output data (not used for our experiments): inlet gas volume fraction ($GVFin$), outlet gas volume fraction ($GVFout$), inlet massflow ($Qin$), outlet massflow ($Qout$). Surge Distance is computed from inlet massflow using the following equations: $ SD=100\, \frac{\phi - \phi_{surge}} {\phi_{surge}} $ with ${\phi_{surge}} = 0.076$ and $ \phi = \frac{Q_{in}}{2.93\, 10^{-3}\, N_{speed}} $. Examples of input and output data is shown on figure \ref{fig:data}. Input variables are cross correlated except $Tin$ and $Pin$.
Output variables are not correlated with $Tin$ and $Pin$ and only slightly correlated with $N$, $\Delta$P and $Power$.

We expect Surge distance to be as well predicted as inlet Massflow given that it can be derived from it by an algebraic equation. Yet, because we use the difference of two close numbers, accuracy requirements become much more challenging for surge distance than it is for inlet massflow.

%

Dataset contains 63116 samples. We use the full dataset for model selection (80\% train, 20\% test and validation, see \ref{sec:model_selection}).
Only a limited number of samples is used for training the active learning model because in real environment we cannot leverage in such large dataset.

\section{Model}

We build three models to find the best architecture for this dataset. These models require a big amount of data to find a good model and in real environment we cannot leverage in such large dataset. For this reason, we developed an active learning method using only a limited number of samples and compare it with the best architecture. The active learning method presented in this paper requires considerably less data to find an accurate estimation model.

\subsection{Model selection}
\label{sec:model_selection}
\small
We apply XGBoost \cite{Chen_2016} to test a boosting mechanism as baseline. Because data on this project are bound to be timeseries, the second model that we developed was Long Short-Term Memory (LSTM) \cite{LSTM} with one layer. Indeed, we've been working on synthetic data coming from a DOE but in the future, data will be on site measurements. The third model that we developed is a multilayer perceptron (MLP) with two layers as an intermediate model between XGBoost and LSTM (see Appendix \ref{app:models} for further details). 

\subsection{Active Learning}
\normalsize
We apply various regression models on this dataset. The results analysis of residuals show heteroscedasticity \cite{Wakefield_2013}. 
We also notice that all points in the dataset may not be useful for the prediction of the output. Thus we explore strategies and efficient approaches to select data points for which MSE can be minimized. We let algorithm choose the potential data points. 

\begin{figure}[htbp]
\centerline{\includegraphics[width=0.5\textwidth]{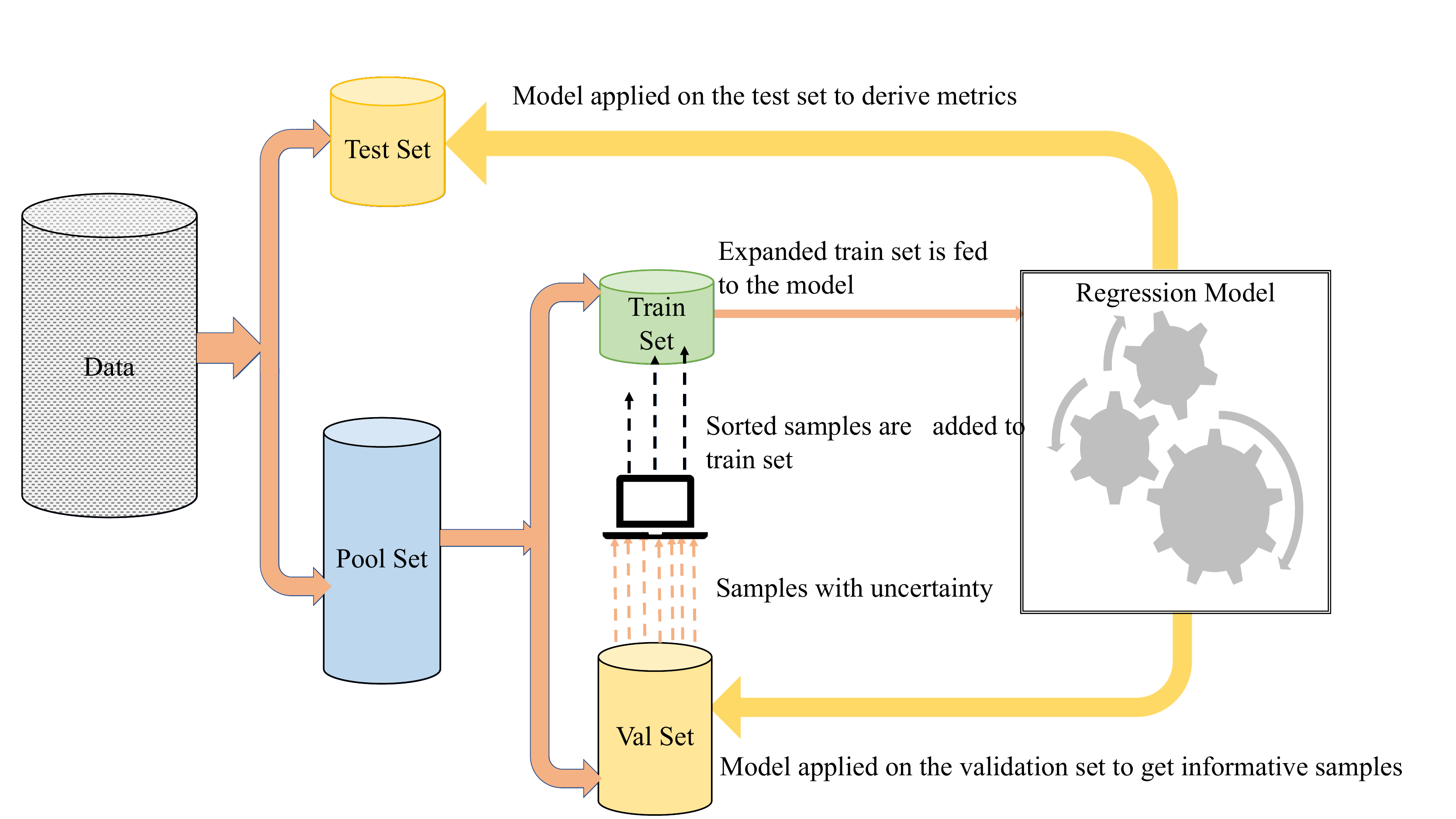}}
\caption{Active Learning Process}
\label{fig:AL_Architecture}
\end{figure}

Given that most of the active learning research happens in the direction of pool based approach, we use this method (Figure \ref{fig:AL_Architecture}). It helps extracting samples based on information measure. Pool based technologies reduce the dimension of data while selecting those that have most information. The idea is to rank the samples based on criteria chosen by the model and  query the most effective data points to the model when building it \ref{fig:AL_Architecture}).
We use the Actively learned model (AL) library \cite{Ghosh_2021} UQ360. 
This AL model includes an ensemble of Neural Networks for training and predictions 
which increases its efficiency \cite{Lakshminarayanan_2016, thebelt2022multi}. 
We use five Neural Networks on the training data with different random batches on each training iteration \cite{goodfellow2016deep} for this experiment. 
We implement the neural network ensemble using pytorch \cite{pytorch} on a GPU machine NVIDIA DGX-1. It is made of an input layer of five nodes, three hidden layers with 256 relu and one last layer which consists of a scaled hyperbolic tangent activation function and one unit with a softplus activation function \cite{goodfellow2016deep}. 
The algorithm returns the mean and log variance (Algorithm \ref{alg:al}). The cost function is negative-log-likelihood of a gaussian function. The mean and variance of the ensemble are the pooled mean and variance. The optimizer is Adam. The learning rate is 0.001 and after the tenth epoch, the learning rate is decayed by a factor of 0.99.

\section{Results}
\subsection{Model Selection}
\label{sec:results}
We compute the scores for the prediction of the surge distance $SD$. To train the model we use 50492 samples and we test it using the remaining 12624 samples.
Based on our previous experiences in energy pump design and client requests, we define as acceptance criteria that the error of $SD$ prediction should be $< \pm 4\%$. We define different metrics to better analyze our results. $R^2$ is convenient because it is independent of the order of magnitude of the variable. We quickly see if it is close to 1. Yet, even a seemingly very good $R^2$ can lead to results that don't reach our acceptance criteria. Therefore, we also take a look at RMSE which is a standard metrics. Max Error is useful to have a feeling of how bad our model can do. We really look for the worst case with this metrics. Finally, we check the Mean Absolute Percentage Error (MAPE), because our criteria is bound to a relative error. At first, we also checked the Maximum Percentage Error, but this one becomes way too large when surge distance is close to 0. As a general fact, relative errors are cumbersome because they can become arbitrary large. At the end, because we don't reach the expected +/- 4\% error for all the data, we check the amount of them reaching this goal to figure out which model is the best one.



Table \ref{tab:errors} reports errors metric. 

\begin{table}[htbp]
\begin{center}
\begin{tabular}{ l | c  | c | c}

Model & LSTM & XGBoost & MLP  \\
\hline
Max Error & 24.8694 & 29.3937 & 29.9472 \\
RMSE & 2.6096 & 4.8438 & 2.8075 \\
$R^2$ & 0.9832 & 0.9534 & 0.9843 \\
MAPE & 0.6725 & 3.2108 & 0.4468 \\
\hline
Acceptance Criteria \\Accuracy  & 52.5\% & 49.7\% & 61.3\%\\  

\hline
\end{tabular}
\caption{LSTM, XGBoost and MLP model errors.}
\label{tab:errors}
\end{center}
\end{table}

Overall, we see that \textbf{XGBoost model} is doing sensitively worst on Rsquared, RMSE and MAPE metrics. Indeed, this is the simplest model and we expected it to be less accurate than the others. \textbf{LSTM model} is slightly better than \textbf{MLP model} for RMSE, equivalent for Rsquared and slightly worst for MAPE. These metrics are not enough to disentangle these models. The most important evaluation is on the acceptance criteria defined before. For \textbf{XGBoost model}, we achieve $\pm$ 4\% of error for 49.7\% of data for Surge Distance.
For \textbf{LSTM model}, we achieve $\pm$ 4\% of error for 52.5\% of data for Surge Distance ($SD$). For \textbf{MLP model}, we achieve $\pm$ 4\% of error for 61.3\% of data for Surge Distance. XGBoost and LSTM give about the same level of accuracy for surge distance, while the MLP achieves higher accuracy. Figure \ref{fig:NN_predicted_SD} shows predicted vs ground truth. With these results, we already tend to prefer \textbf{MLP model}. We'll now check how these models behave with a reduced number of samples before we make a final choice on the model. We train the XGB, MLP and LSTM models on data chunks of similar size as for AL expt. We then apply them on test data and show results in the following table. From the table we see that MLP model performs better than the other two.

\begin{table}[htbp]
\begin{center}
\begin{tabular}{ l | c c c c c c c c   }

\textbf{Partition} & 50	&100	&250	&500 & 750	&1000	&1500	&2000\\
\hline

\textbf{XGBoost} & 8.0984	&8.5688	&7.7655 & 7.0576 & 4.9501	&5.0289	&5.6088	&6.1381 \\
\textbf{MLP} & 1.9839&	1.8783&	2.2285&	1.9265 & 1.9245	&1.8308	&1.8441&	1.8902 \\ 
\textbf{LSTM} & 6.8740	&6.8332	&6.6250	&6.2936 & 5.7186&	5.6452&	5.5331&	5.1076 \\ 
\end{tabular}
\caption{XGBoost, MLP and LSTM errors to predict surge distance with reduced training dataset.}
\label{tab:errors_xgb_mlp_lstm}
\end{center}
\end{table}

Given these results, we choose MLP to predict surge distance for active learning experiments. In addition, we use an ensemble of MLPs to improve results.

\subsection{Active learning}
 
\begin{table}[htbp]
\begin{center}
\begin{tabular}{ l | c c c c c c c c   }

\textbf{Partition} & 1 & 2 & 3 & 4 & 5 & 6 & 7 & 8\\
\hline

\textbf{MLP} & 0.99 & 0.98 & 0.98 & 0.97 & 0.95 & 0.90 & 0.80 & 0.65  \\
\textbf{Active learning} & \textbf{0.93} & \textbf{0.92} & \textbf{0.91} & \textbf{0.89}  & \textbf{0.83} & \textbf{0.73} & \textbf{0.60} & \textbf{0.56}\\ 
\end{tabular}
\caption{Active Learning and MLP errors to predict surge distance.}
\label{tab:errors_al}
\end{center}
\end{table}

To assess AL as a valuable method for real application (when we cannot test thousands of configurations), we select a limited number of samples from our dataset (n = 2200 samples).

\begin{figure}
     \centering
     \begin{subfigure}[b]{0.45\textwidth}
         \centering
         \includegraphics[width=\textwidth]{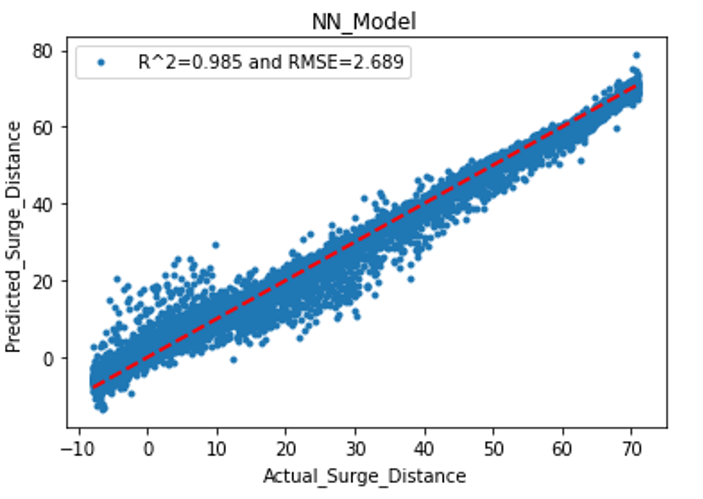}
         \caption{MLP prediction of Surge Distance. Predicted values are shown on the $y$ axis with respect to the ground truth $x$.}
         \label{fig:NN_predicted_SD}
     \end{subfigure}
     \hfill
     \begin{subfigure}[b]{0.45\textwidth}
         \centering
         \includegraphics[width=\textwidth]{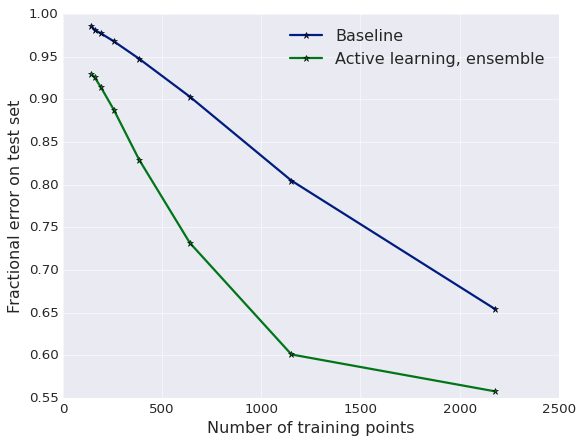}
         \caption{Active Learning model results. AL achieves good performances even after only 1000 samples used for training. }
         \label{fig:AL_results}
     \end{subfigure}
        \caption{Results}
        \label{fig:results}
\end{figure}

We compare the performance of the baseline MLP model with active learning MLP for the same set of data points and the results derived out of the active learning model are promising (see Table \ref{tab:errors_al}). Figure \ref{fig:AL_results} shows the performance of the model trained with active learning and baseline (random selection of train sample). Test set is the same for both cases. AL achieves good performances even after only 1000 samples used for training. Note (although not shown in the figure) that the simple MLP model achieves the performance of AL using about 3000 samples, which is not acceptable for our use case. Finally, we could face issues when going from synthetic data to measured ones. In that case, we would need to retrain the model for each machine and active learning would help us reduce the cost of model retraining by limiting the amount of necessary data.

\section{Conclusion}
\small
We apply active learning to explore the capability of building a surrogate model and improving our design capability. We want to build a surrogate model in order to reduce the time to perform design parameter optimization. Acquiring data from a real test campaign is expensive and costly, therefore we design an active learning framework to train the most accurate model possible with as few data points as possible. Results are very promising suggesting that active learning can be used to define design of experiment for energy applications in an efficient and less expensive way.


\bibliography{biblio}

\appendix

\section*{Appendix A - Symbols}
\begin{table}[htbp]
\begin{center}
\begin{tabular}{ l r r }

Variable & Description & I/O \\
\hline
$Tin$ & inlet temperature & input \\
$Pin$ & inlet pressure & input \\
$N_{speed}$ & rotational speed & input \\
$\Delta$P = Pin - Pout & differential pressure & input \\
$Power$ & total consumed power & input \\
\hline

$SD$ & surge distance & output \\
$Qin$ & inlet massflow & out of scope \\
$GVFin$ & inlet gas volume fraction & out of scope\\
$GVFout$ & outlet gas volume fraction & out of scope\\
$Qout$ & outlet massflow & out of scope

\end{tabular}
\caption{Measures: input (top) and output (bottom).}
\label{tab:symbols}
\end{center}
\end{table}

\section*{Appendix B - Algorithm}
\label{app:algorithm}

We want to compare the performance of the AL model with a baseline. In the baseline, batches of data are sampled randomly while they are selected with the higher variance and
thus with more information in the case of AL approach. Number of samples, configuration and number of iterations are the same for AL and baseline models.
Baseline model gives the predictions from which error is calculated. AL model outputs error estimate which is a measure of uncertainty and in this case variance, apart from the predictions \cite{Lakshminarayanan_2016}.

For the AL model, we add points with most information together with their labels and remove them from the pool set. As training data set is increased, model is re-trained. This process is repeated with the given number of iterations (Algorithm \ref{alg:al}). We use the same datasets as those explored previously when we selected the underlying model.


\begin{algorithm}
	\caption{Active Learning Model Algorithm}
	\label{alg:al}
	\begin{algorithmic}[1]
    \State Input: Initial set of points(T), number of iterations, M,K
    \State Output: The regression model, predictions, variances
    \State Partition the data into pool and test sets.
    \State Divide the pool data into train set which has initial set of points (T), and validation set where size of the train set $<<$ size of the validation set
    
    \State Train the model E on the initial training set T
        \For {$i=1,2,\ldots$}
             \State  $R_i$ = choose M $\times$  K points from the pool
             \State  Compute the error measures using E
             \State  Select K points in $R_i$  which have highest variance 
             \State  Add those K points to the existing train set  
             \State Train the model $E_i=1$ on this expanded training set $T_i=1$
             \State Remove these K points from the pool
        \EndFor
	\end{algorithmic} 
\end{algorithm}

\section*{Appendix C - Model Details}
\label{app:models}
For \textbf{XGBoost}, we run a random search to tune the hyper-parameters for the Surge Distance prediction with learning rate = 0.1. 
We use as boosting parameter gbtree and as standard evaluation metrics RMSE. The resulting optimal model consists of 700 estimators and has a maximum depth of 8 with minimum child weight of 9. For the boosting hyperparameters we choose a step size shrinkage eta of 0.1, a minimum loss reduction gamma of 0.3. Subsample parameter is set to 0.6 which helps prevent overfitting and balances the relatively large value of maximum depth which instead could ease overfitting. Column subsampling by tree is set to 1.0 which is also the default. And the learning task objective is the regression with squared loss.
For \textbf{Long Short-Term Memory (LSTM)} model, we run partial grid searches to tune the hyperparameters for the Surge Distance prediction. Some parameters such as kernel and recurrent regularizers are fixed without the need to check all combinations with others because we quickly see that value 0 gives much better results than any other one. LSTM model consists of one LSTM layer and three dense layers. They contain 32 neurons each. A last dense layer is used for output. L1-L2 regularizer as well as kernel and recurrent regularizers are set to 0. We use a relatively small learning rate of 0.0005 with 5000 epochs. We apply standard Optimizer Adam, activation is relu for LSTM and dense layers. 
For \textbf{multilayer perceptron (MLP)}, we stick to a relatively basic architecture because we expect our dataset doesn't require a too complex architecture and we want a depth in between XGboost and LSTM.  Hyperparameters tuning is done manually from experience. We use two units made of one dense layer, one dropout layer and one batch normalization layer. In the first unit each layer contains 128 neurons. In the second one, they contain 64 neurons. The third dense layer has just one neuron for output. Dropout magnitude is 5\%. Activation is always relu, optimizer is RMSprop. We use the Mean Absolute Error for the loss and our learning rate is 0.01 with a decay of 0.1.

\section*{Appendix D - Data Sample}
\label{app:data}

\begin{figure*}[htbp]
\centering
\includegraphics[width=\textwidth]{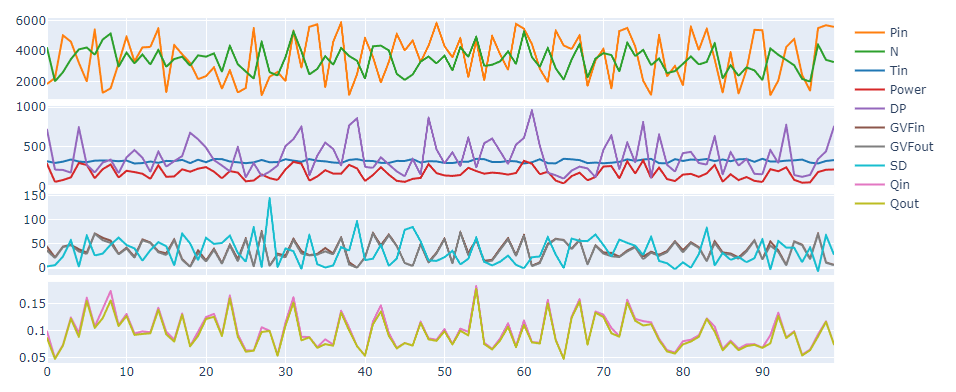}
\caption{Examples of MCP data.}
\label{fig:data}
\end{figure*}

\end{document}